\documentclass[lettersize,journal]{IEEEtran}
\usepackage{amsmath,amsfonts}
\usepackage{algorithmic}
\usepackage{algorithm}
\usepackage{array}
\usepackage[caption=false,font=normalsize,labelfont=sf,textfont=sf]{subfig}
\usepackage{textcomp}
\usepackage{stfloats}
\usepackage{url}
\usepackage{verbatim}
\usepackage{graphicx}
\usepackage{cite}
\begin{document}

\title{SSGaussian: Semantic-Aware and Structure-Preserving\\ 3D Style Transfer}

\author{Jimin Xu, Bosheng Qin, Tao Jin, Zhou Zhao, Zhenhui Ye, Jun Yu, Fei Wu,~\IEEEmembership{Senior Member,~IEEE}
\thanks{The authors are with the College of Computer Science and Technology, Zhejiang University, Hangzhou,
China (e-mail: xujimin@zju.edu.cn).}}

\markboth{Journal of \LaTeX\ Class Files,~Vol.~xx, No.~xx, xx~xxxx}%
{Shell \MakeLowercase{\textit{et al.}}: A Sample Article Using IEEEtran.cls for IEEE Journals}


\maketitle

\begin{abstract}
Recent advancements in neural representations, such as Neural Radiance Fields and 3D Gaussian Splatting, have increased interest in applying style transfer to 3D scenes. While existing methods can transfer style patterns onto 3D-consistent neural representations, they struggle to effectively extract and transfer high-level style semantics from the reference style image. Additionally, the stylized results often lack structural clarity and separation, making it difficult to distinguish between different instances or objects within the 3D scene.
To address these limitations, we propose a novel 3D style transfer pipeline that effectively integrates prior knowledge from pretrained 2D diffusion models. Our pipeline consists of two key stages: First, we leverage diffusion priors to generate stylized renderings of key viewpoints. Then, we transfer the stylized key views onto the 3D representation. This process incorporates two innovative designs.
The first is cross-view style alignment, which inserts cross-view attention into the last upsampling block of the UNet, allowing feature interactions across multiple key views. This ensures that the diffusion model generates stylized key views that maintain both style fidelity and instance-level consistency.
The second is instance-level style transfer, which effectively leverages instance-level consistency across stylized key views and transfers it onto the 3D representation. This results in a more structured, visually coherent, and artistically enriched stylization.
Extensive qualitative and quantitative experiments demonstrate that our 3D style transfer pipeline significantly outperforms state-of-the-art methods across a wide range of scenes, from forward-facing to challenging 360-degree environments. 
Visit our project page \url{https://jm-xu.github.io/SSGaussian/} for immersive visualization.
\end{abstract}

\begin{IEEEkeywords}
3D Style Transfer, 3D Gaussian Splatting, Diffusion.
\end{IEEEkeywords}

\section{Introduction}
\label{sec:intro}

\IEEEPARstart{R}{ecent} advancements in diffusion models have significantly improved image and video generation~\cite{10528891, 10480591, 10589534, 10950092, rombach2022high, ye2023ip}.
In the 3D domain, diffusion models have also been integrated with neural representations, such as Neural Radiance Fields (NeRF)~\cite{mildenhall2021nerf} and 3D Gaussian Splatting (3DGS)~\cite{kerbl20233d}, effectively lifting 2D priors into the 3D world.
DreamFusion~\cite{poole2022dreamfusion} and its follow-up works~\cite{qian2023magic123, melas2023realfusion} introduce Score Distillation Sampling (SDS) loss, which is based on probability density distillation. Through SDS loss, 2D prior knowledge can be utilized to guide 3D content generation. Similarly, Instruct-NeRF2NeRF~\cite{haque2023instruct} and related approaches~\cite{chen2024gaussianeditor, yu2024instantstylegaussian} introduce the Iterative Dataset Update (IDU) method, which employs InstructPix2Pix~\cite{brooks2023instructpix2pix} to iteratively edit rendered images while updating the underlying 3D representations.  
These advancements have led to remarkable progress in 3D generation and editing. However, in the field of 3D style transfer, there still lack effective methods that integrate diffusion priors into a systematically designed 3D style transfer pipeline.

Given a reconstructed 3D scene and a reference style image, 3D style transfer aims to convert the 3D scene into the reference style while preserving the original content structure.
Early 3D style transfer methods were based on point cloud representations~\cite{huang2021learning, mu20223d} or mesh representations~\cite{yin20213dstylenet, michel2022text2mesh}. However, these approaches often produce noticeable artifacts when applied to complex real-world scenes, mainly due to imperfections in geometry reconstruction and texture rendering~\cite{zhang2022arf}.
More recently, neural representations such as NeRF~\cite{mildenhall2021nerf} and 3DGS~\cite{kerbl20233d} have gained significant attention in 3D scene reconstruction. As these neural representations have demonstrated superior fidelity and flexibility, they have also been increasingly adopted in 3D style transfer tasks, replacing traditional point cloud and mesh-based approaches. Current 3D style transfer methods based on NeRF and 3DGS can be broadly categorized into feed-forward approaches~\cite{liu2023stylerf, liu2024stylegaussian} and iterative optimization approaches~\cite{zhang2022arf, kovacs2024g}.
However, both approaches struggle to effectively extract and transfer style semantics from the reference image. Furthermore, the stylized results often lack a layered sense of structure, making it difficult to distinguish between different instances or objects within the 3D scene.

To address these limitations, we propose SSGaussian, a 3D style transfer pipeline that effectively exploits 2D diffusion priors. First, we reconstruct a 3D scene representation using Gaussian Grouping~\cite{ye2024gaussian}, which extends 3DGS to jointly reconstruct and segment anything in 3D scenes. Once the reconstruction is complete, we select key viewpoints and render their corresponding RGB images and depth maps.
Given these key view renderings along with a style image prompt, we leverage a pretrained diffusion model to generate stylized outputs. Diffusion models excel at transferring style semantics while preserving content and structural integrity. However, ensuring multi-view consistency remains a significant challenge, as existing 2D diffusion models struggle to maintain coherence across different viewpoints.
Since the consistency of these stylized key views is crucial for high-quality 3D style transfer, we introduce a Cross-View Style Alignment (CVSA) module to address this issue. Instead of enforcing pixel-level 3D consistency, which is inherently difficult, our module focuses on instance-level consistency—ensuring that the same objects or instances across different key views retain a uniform stylization. To achieve this, we incorporate cross-view attention into the last upsampling block of the UNet~\cite{ronneberger2015u}, allowing feature interactions across multiple key views.
Accompanied with our CVSA module, the diffusion model generates stylized key views that maintain both style fidelity and instance-level consistency, establishing a solid foundation for the subsequent 3D style transfer process.

We further introduce a novel 3D Gaussians stylization algorithm built upon group matching, which effectively transfers stylized key views onto the 3D representation. To fully leverage the instance-level consistency across stylized key views, our key insight is to utilize instance segmentation to establish correspondences between local regions in the training view and their counterparts in the key views. Within each matched local region, we perform nearest-neighbor search and minimize the distance between corresponding features.
Specifically, by leveraging the Identity Encoding parameters introduced in Gaussian Grouping~\cite{ye2024gaussian}, we can obtain group identities for different local regions in any given view. Each group corresponds to a distinct instance in the 3D scene. Using these group identities, we match each group in the training view to the group with the same identity in the stylized key views, ensuring that the same instance is consistently associated across different viewpoints.
Building upon this group matching mechanism, we propose an Instance-level Style Transfer (IST) approach that enables localized and semantically coherent stylization.
For every sampled training view, the objective is to minimize the cosine distance between each feature in a group and its closest counterpart in the stylized key views.
By incorporating this approach, our 3D style transfer method could better preserve both high-level style semantics and fine-grained brushstroke details, which facilitates a more structured, visually coherent, and artistically enriched stylization in the final rendered outputs.

We conduct experiments on a variety of scenes, ranging from forward-facing scenes to challenging 360-degree scenes. To ensure a comprehensive comparison, we select a diverse set of style reference images, enabling a thorough evaluation of the stylization performance of our 3D style transfer pipeline against baseline methods.
Additionally, we perform ablation studies on different components of our pipeline to validate their effectiveness. Both qualitative and quantitative results demonstrate that our 3D style transfer pipeline significantly outperforms state-of-the-art methods. 

\begin{figure*}
    \centering
    \includegraphics[width=\linewidth]{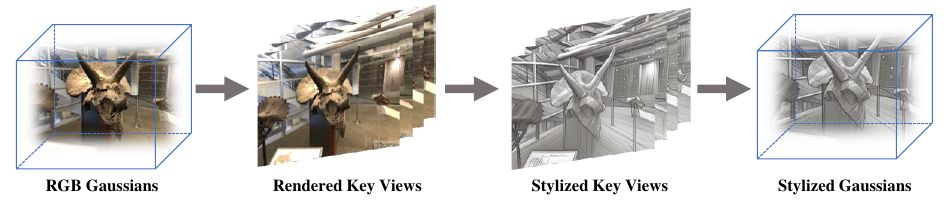}
    \caption{\textbf{Pipeline of SSGaussian.} We begin by reconstructing the scene using a 3D Gaussian Splatting representation. Next, we select key viewpoints and render their corresponding RGB images. Then, given a reference style image, we apply a pretrained diffusion model enhanced with our proposed Cross-View Style Alignment module to generate consistent stylized results for the key views. Finally, we achieve full 3DGS stylization by transferring the stylized key views onto the 3D representation through our Instance-level Style Transfer approach.}
    \label{fig:pipe}
    \vspace{-2mm}
\end{figure*}

We summarize our contributions as follows:
\begin{itemize}

\item We propose a novel 3D style transfer pipeline, which effectively integrates diffusion priors, achieving high-quality and visually coherent style transfer results.
    
\item To address the multi-view consistency challenges of 2D diffusion models, we introduce a Cross-View Style Alignment module, which ensures instance-level consistency across different viewpoints.
    
\item We propose an Instance-level Style Transfer approach with a group matching mechanism, which effectively lifts 2D diffusion priors into 3D stylization.
    
\item Qualitative and quantitative experiments demonstrate that our 3D style transfer pipeline significantly outperforms state-of-the-art methods on a variety of scenes. 
    
\end{itemize}

\section{Related Work}
\label{sec:Related Work}

\begin{figure*}
    \centering
    \includegraphics[width=\linewidth]{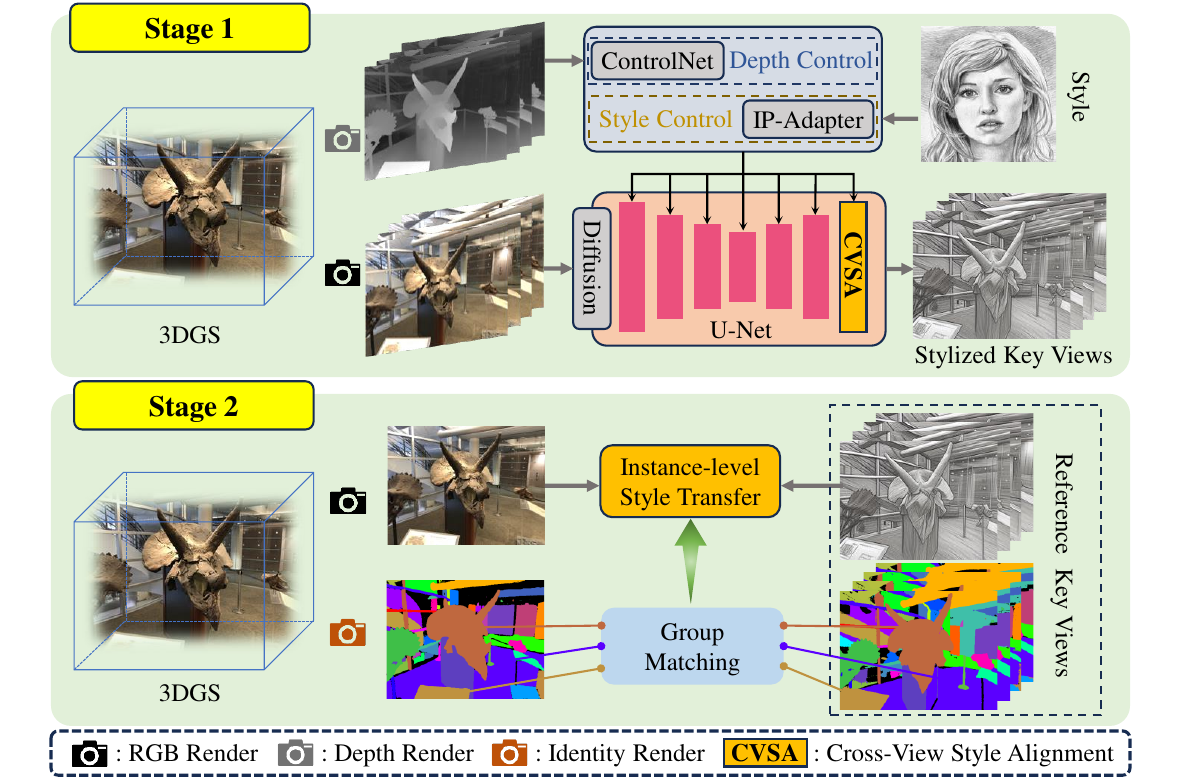}
    \caption{\textbf{Two Stage Stylization.} We decompose the 3D style transfer task into two sequential stages: the stylization of key views and the stylization of the 3D Gaussian Splatting (3DGS) representation based on those stylized key views. In Stage 1, given a style reference image along with RGB and depth images rendered from the 3DGS, we design a diffusion model to effectively transfer style semantics to the selected key viewpoints. In Stage 2, leveraging group matching between the key viewpoints and training views, we introduce an Instance-level Style Transfer approach that hierarchically transfers the style semantics onto the 3DGS representation.}
    \label{fig:alg}
\end{figure*}

\noindent\textbf{Diffusion Model-based Neural Style Transfer.} 
Neural style transfer~\cite{9454281, 9369133, 9525307, 9615003, 10509824, gatys2016image, huang2017arbitrary, jing2019neural} aims at transferring the style of a reference image to the target content image. In the field of text-to-image diffusion models~\cite{rombach2022high}, recent studies have shown that style transfer can be achieved by manipulating the diffusion process through self-attention or cross-attention mechanisms~\cite{sohn2023styledrop, ye2023ip, wang2023styleadapter, hertz2024style, chung2024style, wang2024instantstyle}. 
For example, InstantStyle~\cite{wang2024instantstyle} and Style-Adapter~\cite{wang2023styleadapter} utilize lightweight adapters to extract image features and inject them into cross-attention layers. StyleAlign~\cite{hertz2024style} and StyleID~\cite{chung2024style} perform style transfer by swapping the key and value features of the self-attention block with those from a reference style image.

\noindent\textbf{Text-driven 3D Editing.} 
With the growing popularity of text-to-image diffusion models~\cite{rombach2022high}, researchers are exploring ways to leverage these models for editing 3D scenes based on text instructions. Instruct-NeRF2NeRF (IN2N)~\cite{haque2023instruct} introduces the Iterative Dataset Update optimization algorithm, effectively transforming the 3D NeRF~\cite{mildenhall2021nerf} editing challenge into a 2D image editing task. It then employs InstructPix2Pix (IP2P)~\cite{brooks2023instructpix2pix} to enable instruction-based 2D image editing. ViCA-NeRF~\cite{dong2024vica} offers a more flexible and efficient editing approach by editing key views and incorporating a blending refinement model, ensuring consistent edits without requiring iterative updates in NeRF. DreamEditor~\cite{zhuang2023dreameditor} converts NeRF representations into meshes and directly optimizes the mesh using score distillation sampling (SDS) loss~\cite{poole2022dreamfusion}. 
GaussianEditor~\cite{chen2024gaussianeditor} enhances 3D editing by leveraging the explicit representation properties of Gaussian Splatting~\cite{kerbl20233d}. GaussCtrl~\cite{wu2024gaussctrl} focuses on designing guidance mechanisms within diffusion models, leading to faster editing and improved visual quality.
Other works~\cite{11007035} have explored the use of CLIP~\cite{radford2021learning} for Text-driven 3D Editing.

\noindent\textbf{3D Style Transfer.} 
3D style transfer aims to transform a 3D scene so that renderings from different viewpoints align with the style of a target image while preserving the original content structure. Previous research suggests that stylizing a 3D scene can be explicitly achieved using point cloud~\cite{huang2021learning, mu20223d} and mesh representations~\cite{yin20213dstylenet, michel2022text2mesh}. However, such approaches often produce noticeable artifacts in complex real-world scenes due to imperfections in geometry and texture rendering.~\cite{zhang2022arf} Recently, NeRF~\cite{mildenhall2021nerf} and Gaussian Splatting~\cite{kerbl20233d} have gained significant prominence in 3D scene reconstruction, demonstrating a superior ability to accurately reproduce the appearance of real-world scenes. Consequently, these techniques have also emerged as the dominant 3D representations for 3D style transfer.

One line of approaches follows a feed-forward manner~\cite{liu2023stylerf, liu2024stylegaussian}. These methods require training a neural network for each 3D scene representation, enabling the stylization of the 3D scene in a single forward pass during inference. However, these approaches are limited by the style domain of the training set and often struggle to accurately reproduce fine style patterns, such as colors and brushstrokes. 
Another line of approaches performs 3D style transfer through iterative optimization~\cite{zhang2022arf, pang2023locally, yu2024instantstylegaussian, kovacs2024g}, minimizing content and style losses. ARF~\cite{zhang2022arf} introduces a novel loss function based on nearest neighbor feature matching (NNFM), which better preserves fine details from the style images. LSNeRF~\cite{pang2023locally} proposes a spatial matching mechanism between the style image and NeRF renderings, allowing dynamically assigned correspondences to guide the stylization process and produce diverse results. InstantStyleGaussian~\cite{yu2024instantstylegaussian} incorporates the Iterative Dataset Update optimization algorithm from GaussianEditor~\cite{chen2024gaussianeditor} into 3D style transfer. G-Style~\cite{kovacs2024g} introduces a dual-loss function that enables the algorithm to capture both high-frequency and low-frequency patterns in the style image, enhancing the overall stylization quality.
\section{Method}
\label{sec:Method}

We propose a novel pipeline for 3D style transfer that effectively transfers both large-scale style semantics and fine-scale style patterns (such as brushstrokes), resulting in a more structured and visually coherent stylized rendering. As illustrated in Figure~\ref{fig:pipe}, our pipeline consists of the following key steps: 

\noindent\textbf{Scene Reconstruction.} Reconstruct the scene using a 3D Gaussian Splatting (3DGS)~\cite{kerbl20233d} representation.

\noindent\textbf{Key View Selection and Rendering.} Select key viewpoints and render their corresponding RGB images.

\noindent\textbf{Consistent Multi-view Stylization.} Apply a pretrained diffusion model~\cite{rombach2022high} equipped with our proposed Cross-View Style Alignment module to generate consistent stylized key views.

\noindent\textbf{3DGS Stylization.} Achieve 3DGS stylization with these stylized key views by leveraging our Instance-level Style Transfer approach.

Our 3D style transfer pipeline can be decomposed into two sequential stages: stylization of key views and stylization of the 3DGS representation based on the stylized key views.

\subsection{Preliminaries: 3D Gaussian Grouping}
3D Gaussian Splatting~\cite{kerbl20233d} represents a 3D scene as a collection of 3D colored Gaussians. Each Gaussian is defined by several properties: a mean $\mu \in \mathbb{R}^3$  that determines its center, a covariance matrix $\Sigma \in \mathbb{R}^{3\times3}$ that characterizes its shape and size, an opacity value $\alpha \in \mathbb{R}$, and a color vector $c$. Gaussian Splatting projects these 3D Gaussians onto the 2D image plane, and the color $C$ of a pixel is rendered by blending $N$ ordered Gaussians that overlap with the pixel:
\begin{equation}
\label{eq:rgb}
C=\sum_{i \in N} c_i \alpha_i \prod_{j=1}^{i-1}\left(1-\alpha_j\right).
\end{equation}
Compared to NeRF, 3DGS offers remarkable reconstruction quality with faster rendering speeds. To enable 3D Gaussians for fine-grained scene understanding, Gaussian Grouping~\cite{ye2024gaussian} assigns each Gaussian to its corresponding instances or stuff within the 3D scene. This method first deploys Segment Anything Model (SAM)~\cite{kirillov2023segment} to automatically generate masks for each view independently. It then utilizes a universal temporal propagation model~\cite{cheng2023tracking} to associate mask labels across views, producing a coherent multi-view segmentation as training input. Gaussian Grouping introduces new Identity Encoding parameters, which are integrated into the properties of each 3D Gaussian and jointly optimized during training. The Identity Encoding parameters follow a format similar to color modeling and share a comparable rendering process.

\begin{figure*}
    \centering
    \includegraphics[width=\linewidth]{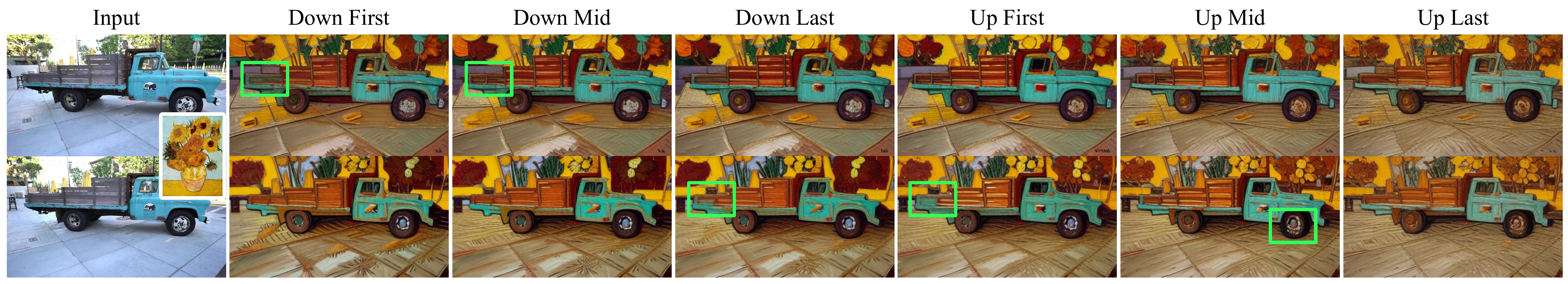}
    \caption{Impact of the Cross-View Style Alignment module across different blocks of the denoising U-Net.}
    \label{fig:ob}
\end{figure*}

\subsection{Consistent Multi-view Stylization.}
As illustrated in Figure~\ref{fig:alg} (Stage 1), we leverage the prior knowledge of a pretrained diffusion model to transfer style semantics to key viewpoints. 
To enable the diffusion model to support style image prompts, we integrate IP-Adapter~\cite{ye2023ip}, which introduces an image encoder to extract features from the style image prompt. A small projection network consisting of a linear layer and Layer Normalization (LN)~\cite{ba2016layer} projects the image embedding into a feature sequence. These features are then embedded into the pretrained diffusion model through adapted modules with decoupled cross-attention.  

To ensure that the generated stylized images preserve the content information of the original views, we employ ControlNet~\cite{zhang2023adding} to regulate the generation process. Benefiting from the 3DGS reconstruction, we can extract consistent depth maps across key views. Using a depth-conditioned ControlNet facilitates the generation of multi-view consistent stylized images.  

Given an input image $I$ to be stylized, a style image prompt $P$, and a depth map $D$, the diffusion model first encodes the input image $I$ into a latent code $z_0$ using a VAE encoder. It then iteratively adds noise to the latent code.  
For this diffusion process, we adopt DDIM inversion~\cite{song2020denoising}, which has been demonstrated by GaussCtrl~\cite{wu2024gaussctrl} to obtain consistent initial noise across multiple views for stable stylization. The DDIM inversion process is formulated as follows: 

\begin{equation}
z_t=\frac{\sqrt{{\alpha_t}}}{\sqrt{{\alpha_{t-1}}}}\left(z_{t-1}-\sqrt{1-\alpha_{t-1}} \epsilon_t\right)+\sqrt{1-{\alpha_t}} \epsilon_t,
\end{equation}
where $t$ is the time step of the diffusion process, $\epsilon_t$ is the noise predicted by the UNet, and $\alpha_t$ is the scheduling coefficient in DDIM scheduler.
After $T$ steps of the diffusion process, we obtain a noisy latent representation $z_T$. The stylized latent representation $z_0'$ is then recovered using the DDIM denoising process: 
\begin{equation}
z_{t-1}'=\sqrt{\alpha_{t-1}}\left(\frac{z_t'-\sqrt{1-{\alpha_t}} \epsilon_t}{\sqrt{{\alpha_t}}}\right)+\sqrt{1-\alpha_{t-1}} \epsilon_t,
\end{equation}

To further ensure instance-level consistency across stylized key views—i.e., maintaining consistent stylization for the same instances or objects across different key views of the 3D scene—we design a Cross-View Style Alignment Module. This module inserts cross-view attention at the last upsampling block of the UNet.   
Before diving into our solution, we revisit the concept of self-attention~\cite{vaswani2017attention} within diffusion models. The self-attention layer takes image features $z \in \mathbb{R}^{HW \times d_{i}}$ as input and computes attention as follows:  
\begin{equation}
  \operatorname{Attn}(z,z)=\operatorname{Softmax}(\frac{Q\left(z\right) K(z)^T}{\sqrt{d}})V(z),
  \label{eq:att}  
\end{equation}
where $Q, K, V$ are linear transformations used to obtain image queries, keys and values $Q\left(z\right) , K(z), V(z) \in \mathbb{R}^{HW \times d}$.

This mechanism enables information exchange across spatial locations within the same view. However, to achieve multi-view consistency, we extend this formulation by introducing cross-view attention, allowing feature interactions across different key views. Specifically, we compute cross-view attention by allowing the query matrix to attend to the key and value matrices of other key views $z^{1:K}$:
\begin{equation}
  \operatorname{Attn}(z,z^{1:K})=\operatorname{Softmax}(\frac{Q\left(z\right) K(z^{1:K})^T}{\sqrt{d}})V(z^{1:K}),
  \label{eq:cvatt}  
\end{equation}

Through empirical analysis, we observe that injecting Cross-View Style Alignment at different blocks of the UNet yields varying degrees of multi-view consistency. As shown in Figure~\ref{fig:ob}, placing the cross-view attention in the early blocks often leads to insufficient semantic alignment across views, such as the material of the truck’s back bucket and tire. In contrast, inserting the module at the last upsampling block consistently achieves the best trade-off between preserving fine-grained style details and maintaining cross-view semantic consistency. We attribute this to the fact that features at the last upsampling stage are semantically rich and spatially refined, making them well-suited for enforcing instance-level consistency across views without disrupting the overall generative fidelity.

This process ensures that the generated stylized images maintain both style fidelity and multi-view consistency, laying a solid foundation for the subsequent 3D style transfer stage.

\begin{figure*}
    \centering
    \includegraphics[width=\linewidth]{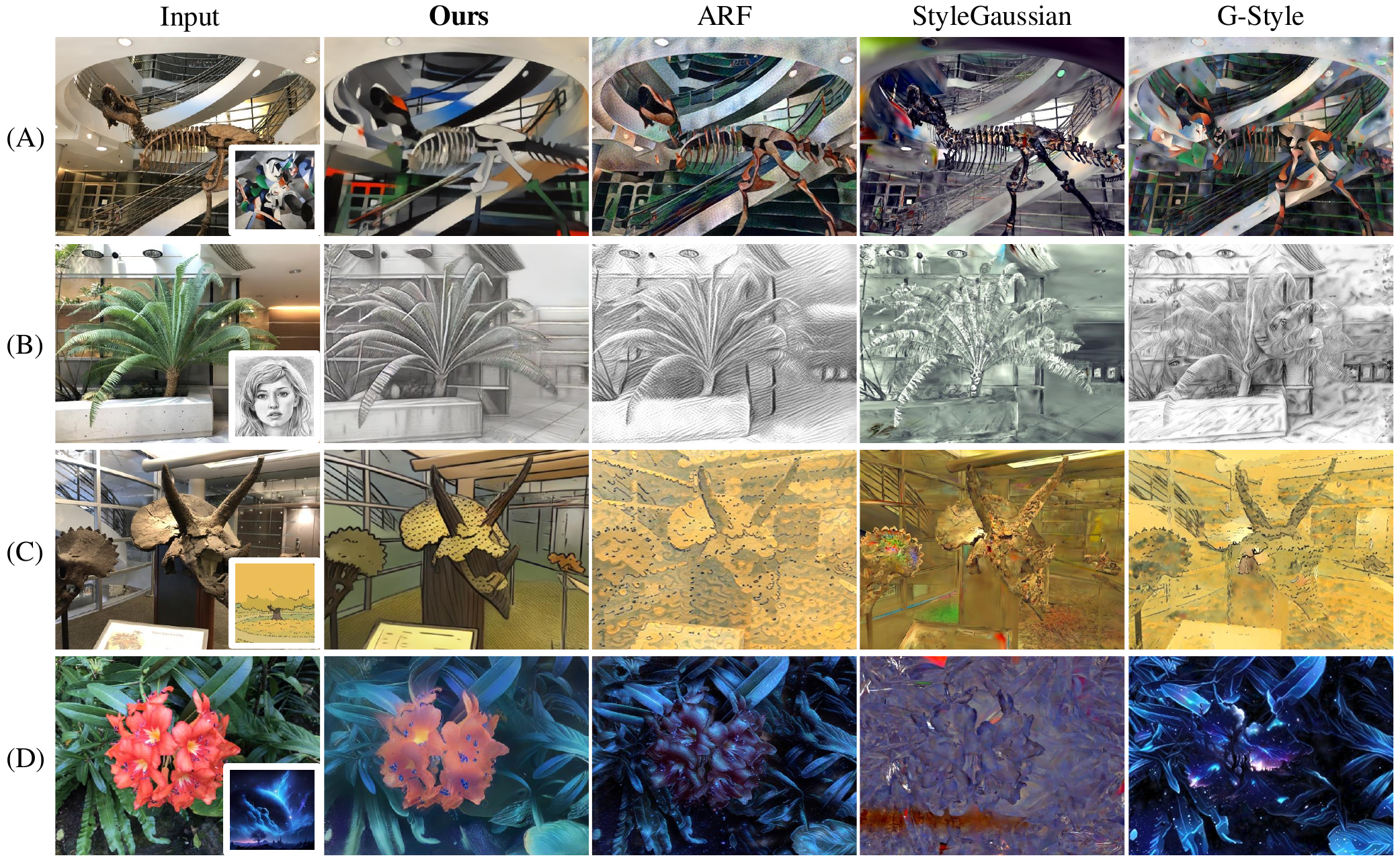}
    \caption{\textbf{Qualitative comparison on LLFF dataset.} We compare our SSGaussian against the state-of-the-art methods on the task of stylizing forward-facing scenes from a reference style image.}
    \label{fig:comparison1}
\end{figure*}

\subsection{3D Gaussians Stylization}
Given the stylized images of key views $S$, our goal is to transfer the stylization to the 3DGS representation, enabling novel view synthesis with consistent stylization.
However, since the stylized key views generated by diffusion model lack strict 3D consistency constraints, they can only ensure instance-level consistency to some extent but do not guarantee pixel-level 3D consistency. As a result, directly fine-tuning 3DGS using these stylized key views often leads to error-prone optimization, where the stylized novel views tend to become blurry and exhibit artifacts.
To better utilize the style information from the key views in guiding the fine-tuning of 3DGS, we propose a 3D Gaussians stylization algorithm build upon a group matching mechanism, as illustrated in Figure~\ref{fig:alg} (Stage 2). This approach enhances the transfer of both style semantics and brushstroke details, leading to a more structured and visually coherent stylization in the final rendered results.

\noindent\textbf{Group Matching}
We first match the local regions of the sampled training view $I$ with those of the stylized key views $S$. Leveraging the Identity Encoding parameters introduced by Gaussian Grouping~\cite{ye2024gaussian}, we can obtain the group identity for different local regions in any given view, where each group represents a specific instance in the 3D scene.
Specifically, Gaussian Grouping introduces new Identity Encoding parameters to each 3D Gaussian. We denote the Identity Encoding as $e_i$, which is a learnable and compact vector of length 16. This encoding is sufficient to distinguish different objects or parts within the scene while maintaining computational efficiency. Similar to equation~\ref{eq:rgb}, we can determine the group identity for each pixel in any given view as follow:
\begin{equation}
\label{eq:id}
E_{id}=\sum_{i \in N} e_i \alpha_i \prod_{j=1}^{i-1}\left(1-\alpha_j\right).
\end{equation}

\begin{figure*}
    \centering
    \includegraphics[width=\linewidth]{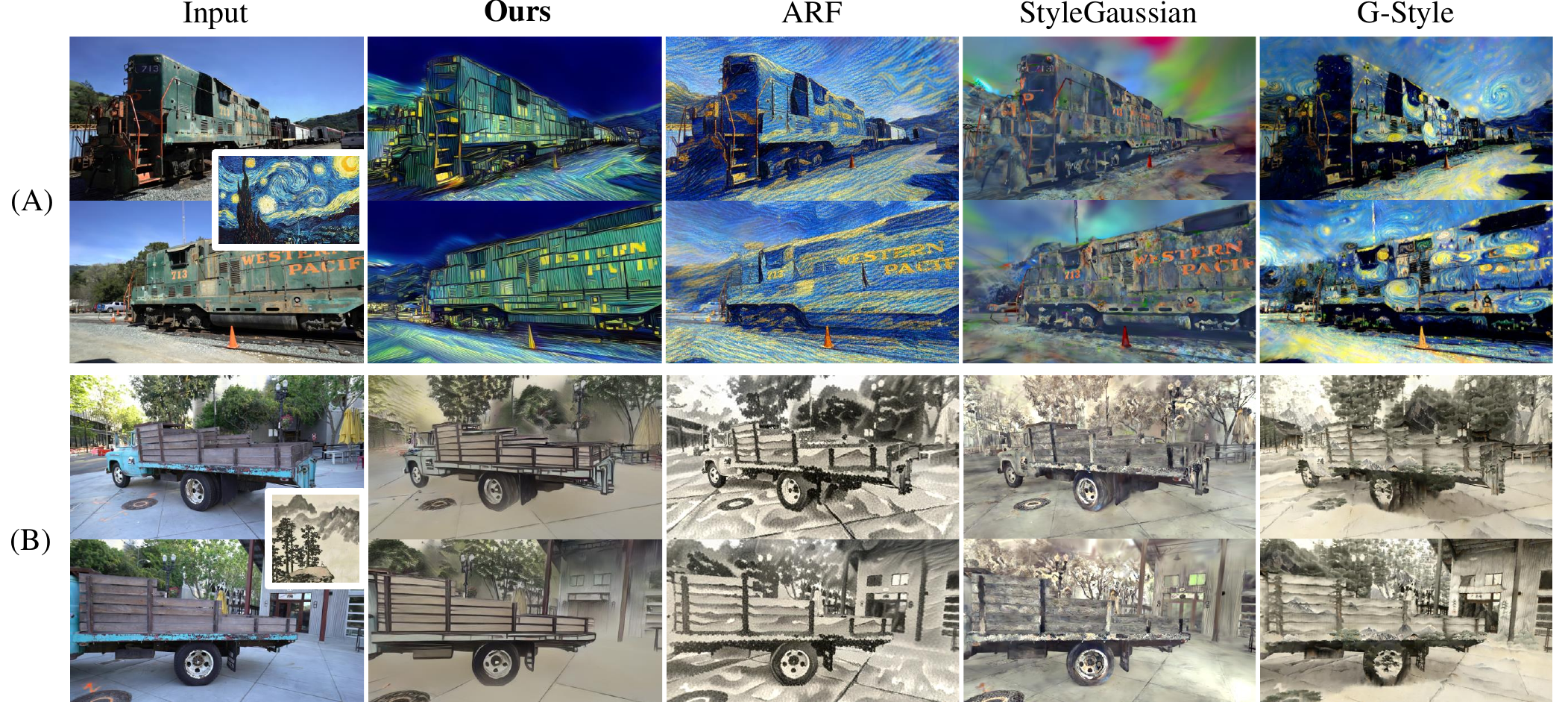}
    \caption{\textbf{Qualitative comparison on Tanks and Temples dataset.} We compare our SSGaussian against the state-of-the-art methods on the task of stylizing 360-degree scenes from a reference style image.}
    \label{fig:comparison2}
\end{figure*}

Next, we apply a linear layer $f$ to restore the feature dimension back to $K$ and then apply a softmax function $softmax(f(E_{id}))$ for identity classification, where $K$ represents the total number of groups in the 3D scene.
Through this process, we obtain the group identity map for any given view. We denote the group identity map of the sampled training view as $M_I$ and that of the stylized key views as $M_S$.
Using the group identity map, we aggregate positions with the same group identity into distinct groups. Let the group sets of the sampled training view and the stylized key views be denoted as $X=\{x_i\}_{i=1}^K$ and $Y=\{y_i\}_{i=1}^K$, respectively.
To provide style guidance for the sampled training view, we match its groups with those of the stylized key views. This matching process can be formulated as constructing a mapping $\mathcal{M}$:
\begin{equation}
  {\mathcal{M}}{(x_i)}=
  \begin{cases}
    y_i, & y_i \neq \emptyset, \\
    \bigcup_{j=1}^{K} y_j, & y_i = \emptyset,
  \end{cases}    
\end{equation}
The mapping matrix $\mathcal{M}$ maps the group with identity $i$ in the sampled training view to the group with the same identity $i$ in the stylized key view, thereby associating the same instance across different views. If there is no region with identity $i$ in the stylized key view, i.e., $y_i = \emptyset$, it is linked to the global region, where the most appropriate style is matched globally.

\noindent\textbf{Instance-level Style Transfer.}
Given the sampled training view $I$, the stylized key views $S$, and their corresponding group matching mapping $\mathcal{M}$, we propose an Instance-level Style Transfer (IST) approach, which performs nearest-neighbor feature matching (NNFM) within local groups.
We extract the high-level VGG feature maps $F_I$ and $F_{S}$ for $I$ and $S$, respectively. Let $F_I{(i, j)}$ denote the feature vector at pixel location $(i, j)$ in the feature map $F_I$. Our style loss is formulated as:
\begin{equation}
\mathcal{L}_S=\frac{1}{N} \sum_{x_k \in X} \sum_{(i, j) \in x_k} \min _{(i', j') \in \mathcal{M}{(x_k)}} D\left(F_I{(i, j)}, F_S{(i', j')}\right),
\end{equation}
where $N$ is the number of pixels in $F_I$, and $D$ computes the cosine distance between two vectors.
Each group in the view can be considered as an instance or stuff within the 3D scene. Our IST approach performs localized nearest-neighbor feature matching between each group in the sampled training view and its corresponding group in the stylized key views. The objective is to minimize the cosine distance between each feature within a group and its nearest neighbor within the feature space of the matching stylized key views.
\section{Experiments}
\label{sec:Experiments}

\subsection{Datasets}
We conduct experiments on two real-world scene datasets: the LLFF dataset~\cite{mildenhall2019local} and the Tanks and Temples dataset~\cite{knapitsch2017tanks}.
The LLFF dataset consists of forward-facing scenes with minimal variation in camera pose. It includes real-world scenes with complex geometric structures; for example, the Fern scene contains highly detailed areas such as leaves, which present a challenge for stylization.
The Tanks and Temples dataset features 360-degree scenes captured in unbounded real-world environments. Unlike LLFF, it includes significant variations in camera poses, making it particularly suitable for evaluating the consistency and robustness of 3D stylization methods.

\subsection{Comparisons with the State of the Arts}
\noindent\textbf{Baselines.}
We compare our approach with three state-of-the-art methods: Artistic Radiance Fields (ARF)~\cite{zhang2022arf}, an iterative optimization approach for NeRF scenes. StyleGaussian~\cite{liu2024stylegaussian}, a feed-forward approach designed for 3DGS scenes. G-Style~\cite{kovacs2024g}, a recent iterative optimization approach applied to 3DGS scenes.

\noindent\textbf{Qualitative comparison.}
We compare our method against three representative baseline approaches on both the LLFF and Tanks and Temples datasets.
In Figure~\ref{fig:comparison1}, we present four forward-facing scenes from the LLFF dataset: Trex, Fern, Horns, and Flower. In Figure~\ref{fig:comparison2}, we showcase two 360-degree scenes from the Tanks and Temples dataset: Train and Truck. For each 360-degree scene, we sample two distant views to display the stylized renderings, allowing for a qualitative comparison of multi-view stylization consistency.
To ensure a comprehensive evaluation, we apply a diverse set of style reference images in Figure~\ref{fig:comparison1} and Figure~\ref{fig:comparison2}, including abstract art, sketches, cartoons, fantasy art, oil paintings, and ink wash paintings. This variety enables a more thorough comparison of the stylization results across different baselines.

From the qualitative comparisons in Figure~\ref{fig:comparison1} and Figure~\ref{fig:comparison2}, we conclude that our method significantly outperforms the baselines in both large-scale style semantics and fine-grained style details.
At a large scale, our approach effectively integrates diffusion model priors, allowing it to incorporate the semantic information from the reference style image while better preserving the content and structure of the original views. For instance, in Figure 4, for highly detailed scenes like Fern and Flower, which contain intricate leaf structures, our 3D style transfer method maintains these fine details more effectively than the baselines. Furthermore, by employing our group matching method, we enable localized stylization for different regions within the 3D scene, leading to a more structured and visually coherent stylization effect. This results in better distinction between different regions, as demonstrated in the stylized Fern and Flower scenes, where the primary subjects are more clearly differentiated.
At a small scale, our method also excels in transferring fine stylistic details such as brushstrokes and object contours more accurately than the baselines. For example, in the Truck scene, the stylized lines of the truck and the brushstroke details of the trees align more closely with the reference style image, further highlighting the superiority of our approach in fine-grained stylization.

\noindent\textbf{Quantitative comparison.}
3D style transfer is a relatively new and under-explored research area. And currently, there still does not exist a standard quantitative metric for evaluating stylization quality. Following previous work~\cite{liu2023stylerf, liu2024stylegaussian}, we focus on assessing multi-view stylization consistency.
Specifically, we measure short-range consistency between two adjacent views, and long-range consistency between two distant views. For any two given views, we warp one view to the other based on optical flow~\cite{teed2020raft} using softmax splatting~\cite{niklaus2020softmax}, and then compute the masked RMSE score and LPIPS score~\cite{zhang2018unreasonable} to quantify the consistency of the stylization.
As shown in Table~\ref{tab:compare}, our approach significantly outperforms baseline methods across both metrics, demonstrating superior stylization consistency. 

Furthermore, to evaluate the quality of stylized renderings, we employ two loss functions commonly used in image style transfer: the Gram Matrix Loss (Style Loss) and the Feature Reconstruction Loss (Content Loss)~\cite{gatys2016image}. Specifically, the Style Loss quantifies the discrepancy between novel views rendered from stylized 3D Gaussians and the style reference image. This is computed as the difference between Gram matrices of feature maps extracted from specific layers of a pretrained VGG network. Lower values indicate greater similarity in style statistics captured at the selected feature layers.
The Content Loss measures the mean squared error (MSE) between feature maps of stylized novel views and their original (non-stylized) counterparts at corresponding viewpoints, again using a pretrained VGG network. Lower MSE values correspond to better preservation of content structure information in the chosen feature layers.
As shown in Table~\ref{tab:compare2}, our approach outperforms baseline methods across both metrics, demonstrating superior stylization quality.

\begin{table}
\caption{\textbf{Quantitative Results.} We evaluate the performance of SSGaussian against state-of-the-art methods in terms of short-range consistency and long-range consistency, using LPIPS (\(\downarrow\)) and RMSE (\(\downarrow\)).}

\centering

\resizebox{\linewidth}{!}{
    \begin{tabular}{*5c}
    \hline
    Method & \multicolumn{2}{c}{\begin{tabular}{@{}c@{}}Short-range \\ Consistency\end{tabular}} & \multicolumn{2}{c}{\begin{tabular}{@{}c@{}}Long-range \\ Consistency\end{tabular}} \\
    \hline
    {}    & \textbf{LPIPS} & \textbf{RMSE}  & \textbf{LPIPS} & \textbf{RMSE} \\
    ARF~\cite{zhang2022arf}       &  0.049  & 0.041  & 0.128  & 0.082   \\
    StyleGaussian~\cite{liu2024stylegaussian}    &  0.036  & 0.030  & 0.077 & 0.071   \\
    G-Style~\cite{kovacs2024g} &  0.035  & 0.035  & 0.089  & 0.072    \\
    \textbf{Ours}                   &  \textbf{0.031}  & \textbf{0.028}  & \textbf{0.073}  & \textbf{0.068}  \\
    \hline
    \end{tabular}
}

\label{tab:compare}
\vspace{-4mm}
\end{table}

\begin{table}
\caption{\textbf{Quantitative Results.} We evaluate the performance of SSGaussian against state-of-the-art methods in terms of renderings quality, using Content Loss (\(\downarrow\)) and Style Loss (\(\downarrow\)).}

\centering

\resizebox{\linewidth}{!}{
    \begin{tabular}{*3c}
    \hline
    Method    & Content Loss & Style Loss \\
    \hline
    ARF~\cite{zhang2022arf}       &  2.490  & 3.184 \\
    StyleGaussian~\cite{liu2024stylegaussian}    &  2.300  & 5.297 \\
    G-Style~\cite{kovacs2024g} &  2.467  & 3.303 \\
    \textbf{Ours}                   &  \textbf{2.298}  & \textbf{3.091} \\
    \hline
    \end{tabular}
}

\label{tab:compare2}
\vspace{-4mm}
\end{table}

\noindent\textbf{Speed comparison.} We measured the training duration and rendering speed of all compared methods using a single NVIDIA RTX 3090 GPU. As shown in Table~\ref{tab:compare3}, our approach achieves efficient stylization and real-time rendering performance comparable to the fastest alternative.
Our approach, along with ARF~\cite{zhang2022arf} and G-Style~\cite{kovacs2024g}, employs iterative optimization for 3D style transfer. Specifically, during each training iteration, these methods directly optimize the 3D scene parameters based on the reference style image. In contrast, StyleGaussian~\cite{liu2024stylegaussian} adopts a feed-forward approach that requires training a dedicated 3D CNN decoder for each 3D scene representation. This enables single-pass stylization during inference.
For the evaluated scenes: Our method completes consistent multi-view stylization in 1 minute and 3D Gaussian stylization in 19 minutes, achieving a rendering speed of 118 FPS. ARF~\cite{zhang2022arf} requires 24 minutes for stylization and, being NeRF-based, renders at 10 FPS. G-Style~\cite{kovacs2024g} consumes 9 minutes in preprocessing and 22 minutes in stylization, rendering at 110 FPS. StyleGaussian~\cite{liu2024stylegaussian} necessitates approximately 5 hours per scene for CNN decoder training, with style transfer operating at 3 FPS.

\noindent\textbf{User study.} To further assess the quality of our approach, we conduct a user study involving 30 participants. Each participant is presented with stylization results from our method, ARF~\cite{zhang2022arf}, StyleGaussian~\cite{liu2024stylegaussian}, and G-Style~\cite{kovacs2024g} alongside their corresponding style exemplars and original content scenes. Without disclosing methodological information, participants evaluate outputs based on three criteria: 
\begin{itemize}
    \item Structural Integrity: Participants are instructed to select the stylization output that best preserve the core structural elements of the original scenes (e.g., object instances, spatial layout, and edge coherence).
    \item Style Similarity: Participants identify outputs that most faithfully match the artistic attributes of the style exemplars, encompassing low-level features (e.g., brushstroke patterns, color palette) and high-level style semantics.
    \item Visual Quality: Participants evaluate outputs based on overall perceptual excellence, considering artifact minimization (e.g., blurring, distortions), multi-view consistency, and aesthetic appeal.
\end{itemize}

As shown in Table~\ref{tab:compare4}, our approach outperforms all comparative methods across every evaluation dimension, demonstrating superior performance in semantic-aware and structure-preserving 3D style transfer.

\subsection{Comparisons with Video Style Transfer Methods}
\begin{figure*}
    \centering
    \includegraphics[width=\linewidth]{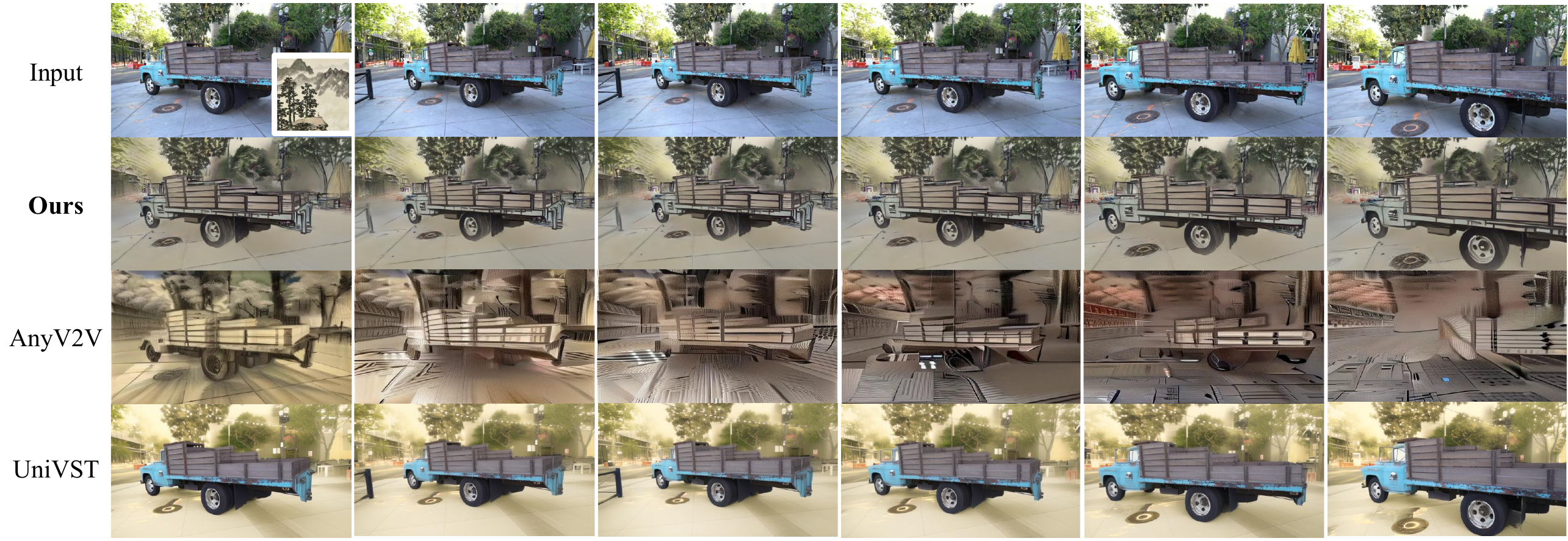}
    \caption{Comparisons with Video Style Transfer Methods.}
    \label{fig:comparisonv}
\end{figure*}

We compare our SSGaussian against video-based style transfer methods AnyV2V~\cite{ku2024anyv2v} and UniVST~\cite{song2024univst}, reformulating the task as temporal stylization by treating multi-view image sequences as video inputs. As evidenced in Figure~\ref{fig:comparisonv}, AnyV2V~\cite{ku2024anyv2v} exhibits temporal inconsistency across frames and structural degradation of scene content. UniVST~\cite{song2024univst} demonstrates inferior style transfer fidelity compared to our approach. SSGaussian achieves superior performance in both style consistency and structural preservation.

\begin{table}
\caption{\textbf{Speed comparison.} 
SSGaussian achieves efficient stylization and real-time rendering performance comparable to the fastest alternative.}

\centering

\resizebox{\linewidth}{!}{
    \begin{tabular}{*3c}
    \hline
    Method    & Training Time & Rendering Speed \\
    \hline
    ARF~\cite{zhang2022arf}       &  24 mins  &  10 FPS  \\
    StyleGaussian~\cite{liu2024stylegaussian}    &  5 hours  &  3 FPS  \\
    G-Style~\cite{kovacs2024g} &  31 mins  &  110 FPS  \\
    \textbf{Ours}                   &  20 mins  &  118 FPS  \\
    \hline
    \end{tabular}
}

\label{tab:compare3}

\end{table}

\begin{table}
\caption{\textbf{User study.} 
The reported values indicate the percentage preference for each method.}

\centering

\resizebox{\linewidth}{!}{
    \begin{tabular}{*4c}
    \hline
    Method    & \multicolumn{1}{c}{\begin{tabular}{@{}c@{}}Structural \\ Integrity\end{tabular}}  & \multicolumn{1}{c}{\begin{tabular}{@{}c@{}}Style \\ Similarity\end{tabular}} & \multicolumn{1}{c}{\begin{tabular}{@{}c@{}}Visual \\ Quality\end{tabular}} \\
    \hline
    ARF~\cite{zhang2022arf}       &  30.0 \%  &  20.0 \%  &  26.7 \%  \\
    StyleGaussian~\cite{liu2024stylegaussian}    &  23.3 \%  &  0.0 \%  &  3.3 \%  \\
    G-Style~\cite{kovacs2024g} &  10.0 \%  &  26.7 \%  &  13.3 \%  \\
    \textbf{Ours}                   &  36.7 \%  &  53.3 \%  &  56.7 \%  \\
    \hline
    \end{tabular}
}

\label{tab:compare4}

\end{table}

\begin{figure}[h]
    \centering
    \includegraphics[width=\linewidth]{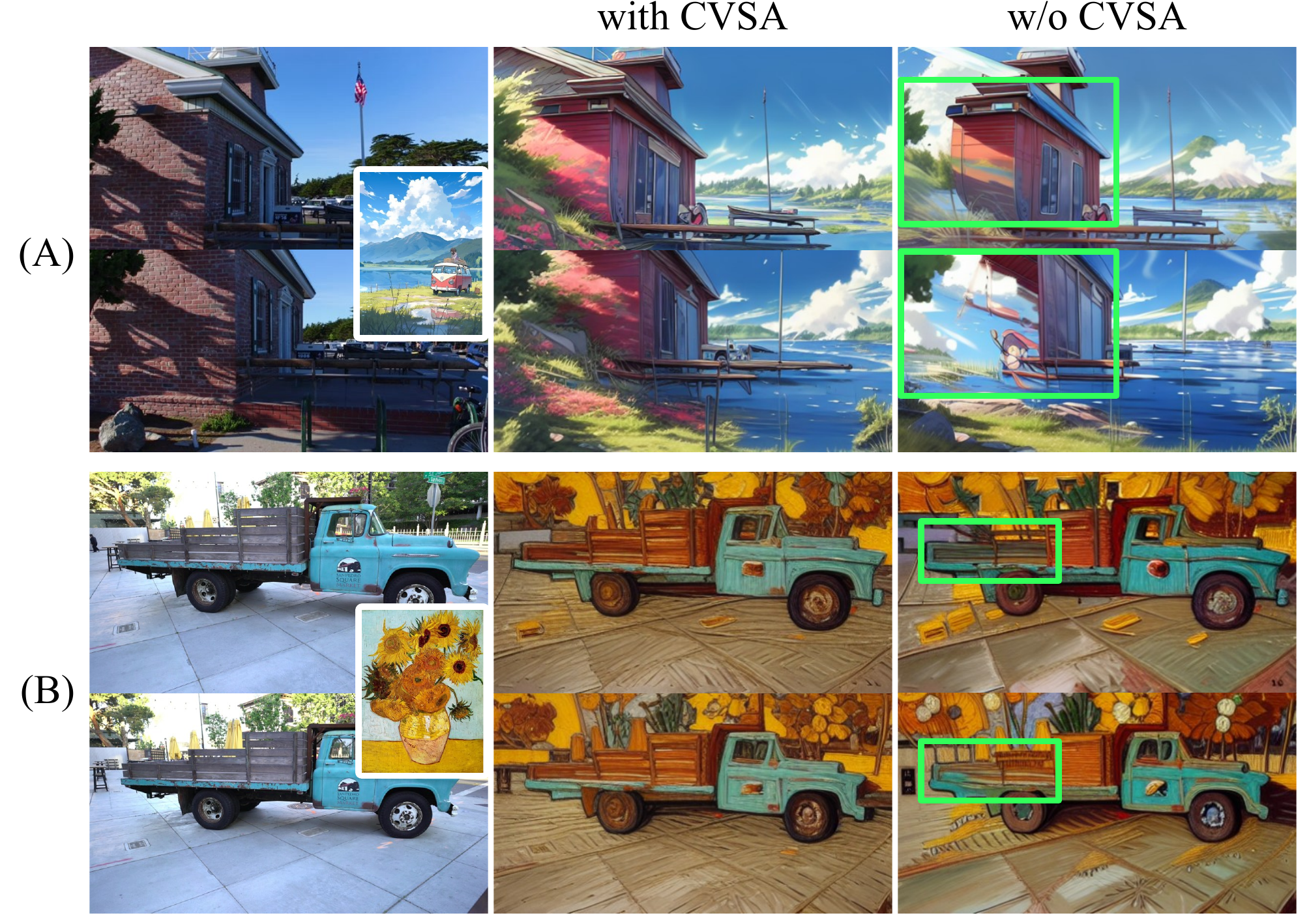}
    \caption{\textbf{Ablation experiment on the proposed Cross-View Style Alignment (CVSA).} (A) and (B) show different scenes, with two views rendered for each. Our CVSA module significantly improves multi-view consistency in both large-scale style semantics and fine-grained style details.}
    \label{fig:ab1}
\end{figure}

\begin{figure}[h]
    \centering
    \includegraphics[width=\linewidth]{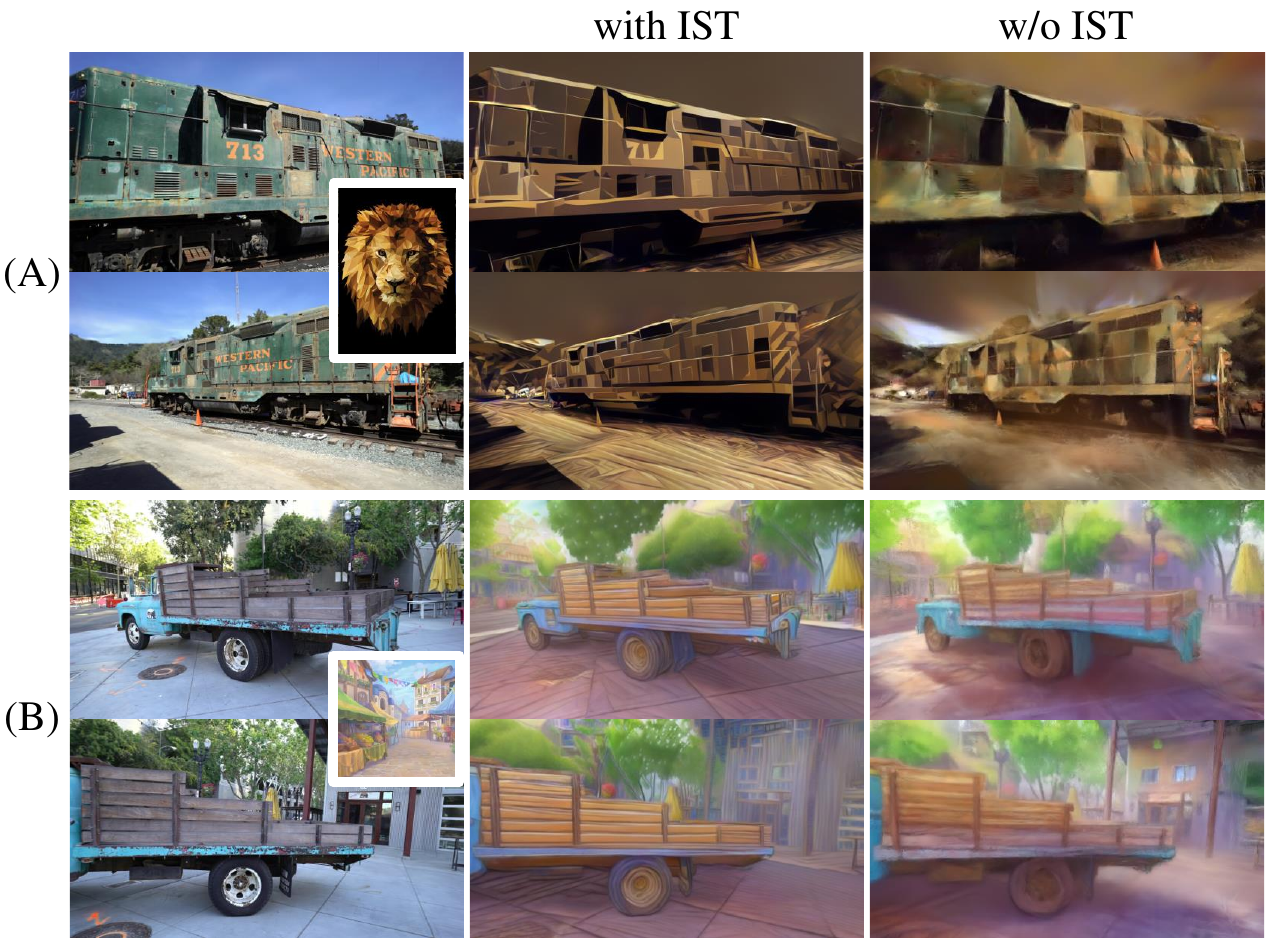}
    \caption{\textbf{Ablation experiment on the proposed Instance-level Style Transfer (IST).} (A) and (B) show different scenes, with two views rendered for each. Our IST approach enables high-quality 3D style transfer by effectively reducing blurriness and mitigating visual artifacts.}
    \label{fig:ab3}
    \vspace{-4mm}
\end{figure}

\subsection{Ablation Studies}
\noindent\textbf{Ablation on Cross-View Style Alignment.}
We perform ablation experiments to evaluate the impact of Cross-View Style Alignment (CVSA) module, as illustrated in Figure~\ref{fig:ab1}. The results demonstrate that our CVSA module significantly improves multi-view consistency in both large-scale style semantics and fine-grained style details. For instance, in Figure~\ref{fig:ab1} (B), CVSA module effectively maintains the color and wooden material of the truck's back bucket. Similarly, in Figure~\ref{fig:ab1} (A), it ensures consistent style transfer across views. Notably, Our CVSA module does not compromise object details: both ablation methods successfully synthesize fine-grained features, such as the label on the truck cabin in Figure~\ref{fig:ab1} (B).

\noindent\textbf{Ablation on Instance-level Style Transfer.}
We conduct an ablation study to evaluate the effectiveness of Instance-level Style Transfer (IST), as illustrated in Figure~\ref{fig:ab3}. We compare our IST approach with the direct fine-tuning approach (w/o IST) in the second stage. In the direct fine-tuning approach, the stylized key views are used to replace the original training views, and the 3DGS is fine-tuned until convergence on these stylized key views. The quality of this baseline heavily depends on the 3D consistency of the stylized key views.  
However, since the stylized key views generated by the diffusion model lack strict 3D consistency constraints, they can only maintain instance-level consistency to some extent but do not guarantee pixel-level 3D consistency. As shown in Figure~\ref{fig:ab3}, this direct fine-tuning approach often results in error-prone optimization, where stylized novel views tend to become blurry and exhibit artifacts. For instance, in Figure~\ref{fig:ab3} (A), artifacts occur at the boundaries of the background. And in Figure~\ref{fig:ab3} (B), the strokes and contours of the truck's back bucket appear blurry.
In contrast, our IST approach achieves high-quality 3D style transfer, effectively transferring strokes and preserving the sharpness of object contours. Additionally, it maintains clear hierarchical distinctions between different instances within the scene, resulting in a more coherent and visually appealing stylization.

\section{Conclusion}
\label{sec:Conclusion}
In this paper, we present a novel 3D style transfer pipeline that effectively integrates prior knowledge from pretrained 2D diffusion models to extract and transfer high-level style semantics from a reference style image. Our pipeline enables the faithful and consistent transfer of these style semantics onto 3D Gaussian Splatting representations. To ensure that the diffusion model generates stylized key views with both style fidelity and instance-level consistency, we introduce a Cross-View Style Alignment module, which incorporates cross-view attention into the last upsampling block of the UNet, facilitating feature interactions across multiple key viewpoints. Additionally, we propose an Instance-level Style Transfer approach, which exploits instance-level consistency across stylized key views and transfers it onto the 3D representation in a semantically structured manner.
Extensive experiments across a wide variety of scenes—including forward-facing and challenging 360-degree environments—demonstrate the superiority of our 3D style transfer pipeline over existing methods.

\bibliographystyle{IEEEtran}
\bibliography{IEEEabrv,reference}

\begin{thebibliography}{10}
\providecommand{\url}[1]{#1}
\csname url@samestyle\endcsname
\providecommand{\newblock}{\relax}
\providecommand{\bibinfo}[2]{#2}
\providecommand{\BIBentrySTDinterwordspacing}{\spaceskip=0pt\relax}
\providecommand{\BIBentryALTinterwordstretchfactor}{4}
\providecommand{\BIBentryALTinterwordspacing}{\spaceskip=\fontdimen2\font plus
\BIBentryALTinterwordstretchfactor\fontdimen3\font minus \fontdimen4\font\relax}
\providecommand{\BIBforeignlanguage}[2]{{%
\expandafter\ifx\csname l@#1\endcsname\relax
\typeout{** WARNING: IEEEtran.bst: No hyphenation pattern has been}%
\typeout{** loaded for the language `#1'. Using the pattern for}%
\typeout{** the default language instead.}%
\else
\language=\csname l@#1\endcsname
\fi
#2}}
\providecommand{\BIBdecl}{\relax}
\BIBdecl

\bibitem{10528891}
Y.~Xu, X.~Xu, H.~Gao, and F.~Xiao, ``Sgdm: An adaptive style-guided diffusion model for personalized text to image generation,'' \emph{IEEE Transactions on Multimedia}, vol.~26, pp. 9804--9813, 2024.

\bibitem{10480591}
C.~Zhang, W.~Yang, X.~Li, and H.~Han, ``Mmginpainting: Multi-modality guided image inpainting based on diffusion models,'' \emph{IEEE Transactions on Multimedia}, vol.~26, pp. 8811--8823, 2024.

\bibitem{10589534}
Y.~Jiang, Q.~Liu, D.~Chen, L.~Yuan, and Y.~Fu, ``Animediff: Customized image generation of anime characters using diffusion model,'' \emph{IEEE Transactions on Multimedia}, vol.~26, pp. 10\,559--10\,572, 2024.

\bibitem{10950092}
H.~Chen, X.~Wang, G.~Zeng, Y.~Zhang, Y.~Zhou, F.~Han, Y.~Wu, and W.~Zhu, ``Videodreamer: Customized multi-subject text-to-video generation with disen-mix finetuning on language-video foundation models,'' \emph{IEEE Transactions on Multimedia}, vol.~27, pp. 2875--2885, 2025.

\bibitem{rombach2022high}
R.~Rombach, A.~Blattmann, D.~Lorenz, P.~Esser, and B.~Ommer, ``High-resolution image synthesis with latent diffusion models,'' in \emph{Proceedings of the IEEE/CVF conference on computer vision and pattern recognition}, 2022, pp. 10\,684--10\,695.

\bibitem{ye2023ip}
H.~Ye, J.~Zhang, S.~Liu, X.~Han, and W.~Yang, ``Ip-adapter: Text compatible image prompt adapter for text-to-image diffusion models,'' \emph{arXiv preprint arXiv:2308.06721}, 2023.

\bibitem{mildenhall2021nerf}
B.~Mildenhall, P.~P. Srinivasan, M.~Tancik, J.~T. Barron, R.~Ramamoorthi, and R.~Ng, ``Nerf: Representing scenes as neural radiance fields for view synthesis,'' \emph{Communications of the ACM}, vol.~65, no.~1, pp. 99--106, 2021.

\bibitem{kerbl20233d}
B.~Kerbl, G.~Kopanas, T.~Leimk{\"u}hler, and G.~Drettakis, ``3d gaussian splatting for real-time radiance field rendering.'' \emph{ACM Trans. Graph.}, vol.~42, no.~4, pp. 139--1, 2023.

\bibitem{poole2022dreamfusion}
B.~Poole, A.~Jain, J.~T. Barron, and B.~Mildenhall, ``Dreamfusion: Text-to-3d using 2d diffusion,'' in \emph{The Eleventh International Conference on Learning Representations}, 2023.

\bibitem{qian2023magic123}
G.~Qian, J.~Mai, A.~Hamdi, J.~Ren, A.~Siarohin, B.~Li, H.-Y. Lee, I.~Skorokhodov, P.~Wonka, S.~Tulyakov \emph{et~al.}, ``Magic123: One image to high-quality 3d object generation using both 2d and 3d diffusion priors,'' in \emph{The Twelfth International Conference on Learning Representations}, 2024.

\bibitem{melas2023realfusion}
L.~Melas-Kyriazi, I.~Laina, C.~Rupprecht, and A.~Vedaldi, ``Realfusion: 360deg reconstruction of any object from a single image,'' in \emph{Proceedings of the IEEE/CVF conference on computer vision and pattern recognition}, 2023, pp. 8446--8455.

\bibitem{haque2023instruct}
A.~Haque, M.~Tancik, A.~A. Efros, A.~Holynski, and A.~Kanazawa, ``Instruct-nerf2nerf: Editing 3d scenes with instructions,'' in \emph{Proceedings of the IEEE/CVF International Conference on Computer Vision}, 2023, pp. 19\,740--19\,750.

\bibitem{chen2024gaussianeditor}
Y.~Chen, Z.~Chen, C.~Zhang, F.~Wang, X.~Yang, Y.~Wang, Z.~Cai, L.~Yang, H.~Liu, and G.~Lin, ``Gaussianeditor: Swift and controllable 3d editing with gaussian splatting,'' in \emph{Proceedings of the IEEE/CVF Conference on Computer Vision and Pattern Recognition}, 2024, pp. 21\,476--21\,485.

\bibitem{yu2024instantstylegaussian}
X.-Y. Yu, J.-X. Yu, L.-B. Zhou, Y.~Wei, and L.-L. Ou, ``Instantstylegaussian: Efficient art style transfer with 3d gaussian splatting,'' \emph{arXiv preprint arXiv:2408.04249}, 2024.

\bibitem{brooks2023instructpix2pix}
T.~Brooks, A.~Holynski, and A.~A. Efros, ``Instructpix2pix: Learning to follow image editing instructions,'' in \emph{Proceedings of the IEEE/CVF Conference on Computer Vision and Pattern Recognition}, 2023, pp. 18\,392--18\,402.

\bibitem{huang2021learning}
H.-P. Huang, H.-Y. Tseng, S.~Saini, M.~Singh, and M.-H. Yang, ``Learning to stylize novel views,'' in \emph{Proceedings of the IEEE/CVF International Conference on Computer Vision}, 2021, pp. 13\,869--13\,878.

\bibitem{mu20223d}
F.~Mu, J.~Wang, Y.~Wu, and Y.~Li, ``3d photo stylization: Learning to generate stylized novel views from a single image,'' in \emph{Proceedings of the IEEE/CVF Conference on Computer Vision and Pattern Recognition}, 2022, pp. 16\,273--16\,282.

\bibitem{yin20213dstylenet}
K.~Yin, J.~Gao, M.~Shugrina, S.~Khamis, and S.~Fidler, ``3dstylenet: Creating 3d shapes with geometric and texture style variations,'' in \emph{Proceedings of the IEEE/CVF International Conference on Computer Vision}, 2021, pp. 12\,456--12\,465.

\bibitem{michel2022text2mesh}
O.~Michel, R.~Bar-On, R.~Liu, S.~Benaim, and R.~Hanocka, ``Text2mesh: Text-driven neural stylization for meshes,'' in \emph{Proceedings of the IEEE/CVF Conference on Computer Vision and Pattern Recognition}, 2022, pp. 13\,492--13\,502.

\bibitem{zhang2022arf}
K.~Zhang, N.~Kolkin, S.~Bi, F.~Luan, Z.~Xu, E.~Shechtman, and N.~Snavely, ``Arf: Artistic radiance fields,'' in \emph{European Conference on Computer Vision}.\hskip 1em plus 0.5em minus 0.4em\relax Springer, 2022, pp. 717--733.

\bibitem{liu2023stylerf}
K.~Liu, F.~Zhan, Y.~Chen, J.~Zhang, Y.~Yu, A.~El~Saddik, S.~Lu, and E.~P. Xing, ``Stylerf: Zero-shot 3d style transfer of neural radiance fields,'' in \emph{Proceedings of the IEEE/CVF Conference on Computer Vision and Pattern Recognition}, 2023, pp. 8338--8348.

\bibitem{liu2024stylegaussian}
K.~Liu, F.~Zhan, M.~Xu, C.~Theobalt, L.~Shao, and S.~Lu, ``Stylegaussian: Instant 3d style transfer with gaussian splatting,'' in \emph{SIGGRAPH Asia 2024 Technical Communications}, 2024, pp. 1--4.

\bibitem{kovacs2024g}
{\'A}.~S. Kov{\'a}cs, P.~Hermosilla, and R.~G. Raidou, ``G-style: Stylized gaussian splatting,'' in \emph{Computer Graphics Forum}, vol.~43, no.~7.\hskip 1em plus 0.5em minus 0.4em\relax Wiley Online Library, 2024, p. e15259.

\bibitem{ye2024gaussian}
M.~Ye, M.~Danelljan, F.~Yu, and L.~Ke, ``Gaussian grouping: Segment and edit anything in 3d scenes,'' in \emph{European Conference on Computer Vision}.\hskip 1em plus 0.5em minus 0.4em\relax Springer, 2024, pp. 162--179.

\bibitem{ronneberger2015u}
O.~Ronneberger, P.~Fischer, and T.~Brox, ``U-net: Convolutional networks for biomedical image segmentation,'' in \emph{Medical image computing and computer-assisted intervention--MICCAI 2015: 18th international conference, Munich, Germany, October 5-9, 2015, proceedings, part III 18}.\hskip 1em plus 0.5em minus 0.4em\relax Springer, 2015, pp. 234--241.

\bibitem{9454281}
J.~J. Virtusio, J.~J.~M. Ople, D.~S. Tan, M.~Tanveer, N.~Kumar, and K.-L. Hua, ``Neural style palette: A multimodal and interactive style transfer from a single style image,'' \emph{IEEE Transactions on Multimedia}, vol.~23, pp. 2245--2258, 2021.

\bibitem{9369133}
S.~Liu and T.~Zhu, ``Structure-guided arbitrary style transfer for artistic image and video,'' \emph{IEEE Transactions on Multimedia}, vol.~24, pp. 1299--1312, 2022.

\bibitem{9525307}
H.~Mun, G.-J. Yoon, J.~Song, and S.~M. Yoon, ``Texture preserving photo style transfer network,'' \emph{IEEE Transactions on Multimedia}, vol.~24, pp. 3823--3834, 2022.

\bibitem{9615003}
Y.~Huang, M.~Jing, J.~Zhou, Y.~Liu, and Y.~Fan, ``Lccstyle: Arbitrary style transfer with low computational complexity,'' \emph{IEEE Transactions on Multimedia}, vol.~25, pp. 501--514, 2023.

\bibitem{10509824}
H.~Ding, H.~Zhang, G.~Fu, C.~Jiang, F.~Luo, C.~Xiao, and M.~Xu, ``Towards high-quality photorealistic image style transfer,'' \emph{IEEE Transactions on Multimedia}, vol.~26, pp. 9892--9905, 2024.

\bibitem{gatys2016image}
L.~A. Gatys, A.~S. Ecker, and M.~Bethge, ``Image style transfer using convolutional neural networks,'' in \emph{Proceedings of the IEEE conference on computer vision and pattern recognition}, 2016, pp. 2414--2423.

\bibitem{huang2017arbitrary}
X.~Huang and S.~Belongie, ``Arbitrary style transfer in real-time with adaptive instance normalization,'' in \emph{Proceedings of the IEEE international conference on computer vision}, 2017, pp. 1501--1510.

\bibitem{jing2019neural}
Y.~Jing, Y.~Yang, Z.~Feng, J.~Ye, Y.~Yu, and M.~Song, ``Neural style transfer: A review,'' \emph{IEEE transactions on visualization and computer graphics}, vol.~26, no.~11, pp. 3365--3385, 2019.

\bibitem{sohn2023styledrop}
K.~Sohn, N.~Ruiz, K.~Lee, D.~C. Chin, I.~Blok, H.~Chang, J.~Barber, L.~Jiang, G.~Entis, Y.~Li \emph{et~al.}, ``Styledrop: text-to-image generation in any style,'' in \emph{Proceedings of the 37th International Conference on Neural Information Processing Systems}, 2023, pp. 66\,860--66\,889.

\bibitem{wang2023styleadapter}
Z.~Wang, X.~Wang, L.~Xie, Z.~Qi, Y.~Shan, W.~Wang, and P.~Luo, ``Styleadapter: A unified stylized image generation model,'' \emph{International Journal of Computer Vision}, vol. 133, no.~4, pp. 1894--1911, 2025.

\bibitem{hertz2024style}
A.~Hertz, A.~Voynov, S.~Fruchter, and D.~Cohen-Or, ``Style aligned image generation via shared attention,'' in \emph{Proceedings of the IEEE/CVF Conference on Computer Vision and Pattern Recognition}, 2024, pp. 4775--4785.

\bibitem{chung2024style}
J.~Chung, S.~Hyun, and J.-P. Heo, ``Style injection in diffusion: A training-free approach for adapting large-scale diffusion models for style transfer,'' in \emph{Proceedings of the IEEE/CVF Conference on Computer Vision and Pattern Recognition}, 2024, pp. 8795--8805.

\bibitem{wang2024instantstyle}
H.~Wang, M.~Spinelli, Q.~Wang, X.~Bai, Z.~Qin, and A.~Chen, ``Instantstyle: Free lunch towards style-preserving in text-to-image generation,'' \emph{arXiv preprint arXiv:2404.02733}, 2024.

\bibitem{dong2024vica}
J.~Dong and Y.-X. Wang, ``Vica-nerf: View-consistency-aware 3d editing of neural radiance fields,'' \emph{Advances in Neural Information Processing Systems}, vol.~36, 2024.

\bibitem{zhuang2023dreameditor}
J.~Zhuang, C.~Wang, L.~Lin, L.~Liu, and G.~Li, ``Dreameditor: Text-driven 3d scene editing with neural fields,'' in \emph{SIGGRAPH Asia 2023 Conference Papers}, 2023, pp. 1--10.

\bibitem{wu2024gaussctrl}
J.~Wu, J.-W. Bian, X.~Li, G.~Wang, I.~Reid, P.~Torr, and V.~A. Prisacariu, ``Gaussctrl: Multi-view consistent text-driven 3d gaussian splatting editing,'' in \emph{European Conference on Computer Vision}.\hskip 1em plus 0.5em minus 0.4em\relax Springer, 2024, pp. 55--71.

\bibitem{11007035}
W.~Liang, H.~Xu, W.~Gan, and W.~Kang, ``Zero-shot text-driven dynamic neural radiance fields stylization,'' \emph{IEEE Transactions on Multimedia}, pp. 1--14, 2025.

\bibitem{radford2021learning}
A.~Radford, J.~W. Kim, C.~Hallacy, A.~Ramesh, G.~Goh, S.~Agarwal, G.~Sastry, A.~Askell, P.~Mishkin, J.~Clark \emph{et~al.}, ``Learning transferable visual models from natural language supervision,'' in \emph{International conference on machine learning}.\hskip 1em plus 0.5em minus 0.4em\relax PmLR, 2021, pp. 8748--8763.

\bibitem{pang2023locally}
H.-W. Pang, B.-S. Hua, and S.-K. Yeung, ``Locally stylized neural radiance fields,'' in \emph{2023 IEEE/CVF International Conference on Computer Vision (ICCV)}.\hskip 1em plus 0.5em minus 0.4em\relax IEEE Computer Society, 2023, pp. 307--316.

\bibitem{kirillov2023segment}
A.~Kirillov, E.~Mintun, N.~Ravi, H.~Mao, C.~Rolland, L.~Gustafson, T.~Xiao, S.~Whitehead, A.~C. Berg, W.-Y. Lo \emph{et~al.}, ``Segment anything,'' in \emph{Proceedings of the IEEE/CVF International Conference on Computer Vision}, 2023, pp. 4015--4026.

\bibitem{cheng2023tracking}
H.~K. Cheng, S.~W. Oh, B.~Price, A.~Schwing, and J.-Y. Lee, ``Tracking anything with decoupled video segmentation,'' in \emph{Proceedings of the IEEE/CVF International Conference on Computer Vision}, 2023, pp. 1316--1326.

\bibitem{ba2016layer}
J.~L. Ba, J.~R. Kiros, and G.~E. Hinton, ``Layer normalization,'' \emph{arXiv preprint arXiv:1607.06450}, 2016.

\bibitem{zhang2023adding}
L.~Zhang, A.~Rao, and M.~Agrawala, ``Adding conditional control to text-to-image diffusion models,'' in \emph{Proceedings of the IEEE/CVF international conference on computer vision}, 2023, pp. 3836--3847.

\bibitem{song2020denoising}
J.~Song, C.~Meng, and S.~Ermon, ``Denoising diffusion implicit models,'' in \emph{International Conference on Learning Representations}, 2021.

\bibitem{vaswani2017attention}
A.~Vaswani, N.~Shazeer, N.~Parmar, J.~Uszkoreit, L.~Jones, A.~N. Gomez, {\L}.~Kaiser, and I.~Polosukhin, ``Attention is all you need,'' \emph{Advances in neural information processing systems}, vol.~30, 2017.

\bibitem{mildenhall2019local}
B.~Mildenhall, P.~P. Srinivasan, R.~Ortiz-Cayon, N.~K. Kalantari, R.~Ramamoorthi, R.~Ng, and A.~Kar, ``Local light field fusion: Practical view synthesis with prescriptive sampling guidelines,'' \emph{ACM Transactions on Graphics (ToG)}, vol.~38, no.~4, pp. 1--14, 2019.

\bibitem{knapitsch2017tanks}
A.~Knapitsch, J.~Park, Q.-Y. Zhou, and V.~Koltun, ``Tanks and temples: Benchmarking large-scale scene reconstruction,'' \emph{ACM Transactions on Graphics (ToG)}, vol.~36, no.~4, pp. 1--13, 2017.

\bibitem{teed2020raft}
Z.~Teed and J.~Deng, ``Raft: Recurrent all-pairs field transforms for optical flow,'' in \emph{Computer Vision--ECCV 2020: 16th European Conference, Glasgow, UK, August 23--28, 2020, Proceedings, Part II 16}.\hskip 1em plus 0.5em minus 0.4em\relax Springer, 2020, pp. 402--419.

\bibitem{niklaus2020softmax}
S.~Niklaus and F.~Liu, ``Softmax splatting for video frame interpolation,'' in \emph{Proceedings of the IEEE/CVF conference on computer vision and pattern recognition}, 2020, pp. 5437--5446.

\bibitem{zhang2018unreasonable}
R.~Zhang, P.~Isola, A.~A. Efros, E.~Shechtman, and O.~Wang, ``The unreasonable effectiveness of deep features as a perceptual metric,'' in \emph{Proceedings of the IEEE conference on computer vision and pattern recognition}, 2018, pp. 586--595.

\bibitem{ku2024anyv2v}
M.~Ku, C.~Wei, W.~Ren, H.~Yang, and W.~Chen, ``Anyv2v: A tuning-free framework for any video-to-video editing tasks,'' \emph{Transactions on Machine Learning Research}, 2024.

\bibitem{song2024univst}
Q.~Song, M.~Lin, W.~Zhan, S.~Yan, L.~Cao, and R.~Ji, ``Univst: A unified framework for training-free localized video style transfer,'' \emph{arXiv preprint arXiv:2410.20084}, 2024.

\end{thebibliography}

\end{document}